\documentclass[conference]{IEEEtran}

\IEEEoverridecommandlockouts

\usepackage{cite}
\usepackage{amssymb} 
\usepackage{amsfonts}
\usepackage{amsmath}
\usepackage{amsthm}
\usepackage{graphicx}
\usepackage{textcomp}
\usepackage{xcolor}
\def\BibTeX{{\rm B\kern-.05em{\sc i\kern-.025em b}\kern-.08em
    T\kern-.1667em\lower.7ex\hbox{E}\kern-.125emX}}

\usepackage{hyperref}
\usepackage{multirow}       
\usepackage{booktabs}       
\usepackage{algorithm} 
\usepackage{algpseudocode} 
\usepackage{subcaption}     

\renewcommand{\lll}{\boldsymbol{\lambda}}

\algnewcommand\algorithmicinput{\textbf{Input:}}
\algnewcommand\Input{\item[\algorithmicinput]}
\algnewcommand\algorithmicoutput{\textbf{Output:}}
\algnewcommand\Output{\item[\algorithmicoutput]}

\newcommand{\ttt}{\boldsymbol{\theta}}
\newcommand{\lbold}{\boldsymbol{\lambda}}

\newtheorem{Prop}{Proposition}

\begin{document}
        \title{Constrained deep networks: Lagrangian optimization via Log-barrier extensions}

        \author{\IEEEauthorblockN{Hoel \textsc{Kervadec}}
        \IEEEauthorblockA{\textit{Erasmus MC} \\
        Netherlands \\
       \url{hoel@kervadec.science}
        }
        \and
        \IEEEauthorblockN{Jose \textsc{Dolz}}
        \IEEEauthorblockA{\textit{ÉTS Montréal} \\
        Canada
        }
        \and
        \IEEEauthorblockN{Jing \textsc{Yuan}}
        \IEEEauthorblockA{\textit{Xidian University}\\
        China
        }
        \and
        \IEEEauthorblockN{Christian \textsc{Desrosiers}}
        \IEEEauthorblockA{\textit{ÉTS Montréal} \\
        Canada
        }
        \and
        \IEEEauthorblockN{Eric \textsc{Granger}}
        \IEEEauthorblockA{\textit{ÉTS Montréal} \\
        Canada
        }
        \and
        \IEEEauthorblockN{Ismail \textsc{Ben Ayed}}
        \IEEEauthorblockA{\textit{ÉTS Montréal} \\
        Canada
        }}
        \maketitle

        \begin{abstract}
                This study investigates imposing hard inequality constraints on the outputs of convolutional neural networks (CNN) during training. Several recent works showed that the theoretical and practical advantages of Lagrangian optimization over simple penalties do not materialize in practice when dealing with modern CNNs involving millions of parameters. Therefore, constrained CNNs are typically handled with penalties.
                We propose {\em log-barrier extensions}, which approximate Lagrangian optimization of constrained-CNN problems with a sequence of unconstrained losses. Unlike standard interior-point and log-barrier methods, our formulation does not need an initial feasible solution.
                The proposed extension yields an upper bound on the duality gap---generalizing the result of standard log-barriers---and yielding sub-optimality certificates for feasible solutions.
                While sub-optimality is not guaranteed for non-convex problems, this result shows that log-barrier extensions are a principled way to approximate Lagrangian optimization for constrained CNNs via {\em implicit} dual variables. We report weakly supervised image segmentation experiments, with various constraints, showing that our formulation outperforms substantially the existing constrained-CNN methods, in terms of accuracy, constraint satisfaction and training stability, more so when dealing with a large number of constraints.
        \end{abstract}


        \section{Introduction}
                \label{sec:introduction}

                Imposing prior knowledge in the form of hard constraints on the output of deep convolutional neural networks (CNNs) is useful in a breadth of
                learning and vision problems. For example, in semi- and weakly-supervised learning, structured prediction or multi-task learning, a set of natural prior-knowledge constraints is available. Such additional knowledge may come from domain experts, for example. In semi-supervised image segmentation, for instance, several recent works \cite{kervadec2019constrained,nandwani2019primal,Zhou2019ICCV} showed that imposing domain-specific knowledge on the network's predictions at unlableled data points acts as a powerful regularizer, boosting significantly the performances when the amount of labeled data is limited. Specifically, the authors of \cite{kervadec2019constrained,Zhou2019ICCV} added priors on the sizes of the target regions, achieving good performances with only fractions of labels. Such constraints are highly relevant in medical imaging \cite{Litjens2017}, and can mitigate the lack of full annotations\footnote{In semantic segmentation, full supervision involves annotating all pixels in each training image, a problem amplified when annotations require expert knowledge or involves volumetric data as in medical imaging.}. Similar experimental observations were made in other application areas of semi-supervised learning. For example, in natural language processing, the authors of \cite{nandwani2019primal}, showed that embedding prior-knowledge constraints on unlabled data can yield significant boosts in performances. 3D human pose estimation from a single view \cite{Marquez-Neila2017} is another application where task-specific prior constraints arise naturally, e.g., symmetry constraints encoding that the two arms should have the same length.

                As pointed out in several studies \cite{kervadec2019constrained,Marquez-Neila2017,nandwani2019primal,pathak2015constrained,Ravi2018,Zhou2019ICCV}, imposing hard constraints on modern deep CNNs involving millions of parameters is challenging, even when the constraints are convex with respect to the outputs of the network. In modern deep networks, constraints are commonly handled with {\em penalties} for their simplicity, and despite their well-known limitations. Standard Lagrangian-dual optimization has been largely avoided and, as discussed in \cite{Marquez-Neila2017,pathak2015constrained,Ravi2018}, this might be explained by the computational complexity and stability/convergence issues caused by alternating between stochastic optimization and {\em explicit} dual updates/projections.

                Interior-point and log-barrier methods can approximate Lagrangian optimization by starting from a feasible solution and solving unconstrained problems, while completely avoiding explicit dual steps and projections. Unfortunately, despite their well-established advantages over penalties, such standard log-barriers were not used before in deep CNNs because finding a feasible set of initial network parameters is not trivial, and is itself a challenging constrained-CNN problem.

                We propose {\em log-barrier extensions}, which approximate Lagrangian optimization of constrained-CNN problems with a sequence of unconstrained losses, removing the need for an initial feasible set of network parameters. The extensions yield a duality-gap bound, which generalizes the standard duality-gap result of log-barriers, yielding sub-optimality certificates for feasible solutions in the case of convex losses. While sub-optimality is not guaranteed for non-convex problems, this result shows that log-barrier extensions are a principled way to approximate Lagrangian optimization for constrained CNNs via {\em implicit} dual variables. This addresses the well-known limitations of penalty methods and, at the same time, removes the explicit dual updates of Lagrangian optimization. We report comprehensive weakly supervised segmentation experiments, with various constraints, showing that our formulation outperforms the existing constrained-CNN methods, in terms of accuracy, constraint satisfaction and training stability, more so when dealing with a large number of constraints.

        \section{Method}
                \label{sec:extensions}

                \subsection{Preliminaries}
                        Let $\mathcal{D} = \{I^1,...,I^N\}$ denotes a partially labeled set of $N$ training images, and $S_{\ttt} = \{s_{\ttt}^1,...,s_{\ttt}^N\}$ denotes the associated predicted networks outputs in the form of softmax probabilities, for both unlabeled and labeled data points, with $\ttt$ the neural-network weights. These could be class probabilities or dense pixel-wise probabilities in the case of semantic image segmentation. We address constrained problems of the following general form:
                        \begin{align}
                                \label{Constrained-CNN}
                                \min_{\ttt} \quad &{\cal E}(\ttt)  \\
                                        s.t. \quad &f_1(s^n_{\ttt}) \leq 0, \quad n=1, \dots, N, \nonumber\\
                                        & \dots \nonumber\\
                                        &f_P(s^n_{\ttt}) \leq 0, \quad n=1, \dots, N . \nonumber
                        \end{align}
                        where ${\cal E}(\ttt)$ is some standard loss over the set of labeled data points---e.g., cross-entropy---and each $f_i$ is some differentiable function, which we want to constrain for each data point $n$.
                        Inequality constraints of the general form in Eq. \eqref{Constrained-CNN} can embed very useful prior knowledge on the network's predictions for unlabeled pixels. Assume, for instance, in the case of image segmentation, that we have prior knowledge about the size of the target region (i.e., class) $k$. Such a knowledge can be in the form of lower or upper bounds on region size, which is common in medical image segmentation problems \cite{Gorelick2013,kervadec2019constrained,Niethammer2013}.
                        In this case, $I^n: \Omega  \subset \mathbb{R}^2  \rightarrow \mathbb{R}$ could be a partially labeled or unlabeled image, with $\Omega$
                        the spatial support of the image, and $s_{\ttt}^n \in [0, 1]^{K\times|\Omega|}$ its predicted mask. This matrix contains the softmax probabilities for each pixel $p \in \Omega$ and each class $k$, which we denotes $s_{k,p,\ttt}^n$.
                        A constraint in the form of $f_i(s^n_{\ttt}) = \sum_{p \in \Omega} s_{k,p,\ttt}^n - a \leq 0$ enforces an upper limit $a$ on the size of target region $k$.

                \subsection{Log-barrier extensions}
                        We propose the following unconstrained loss for approximating Lagrangian optimization of constrained problem \eqref{Constrained-CNN}:
                        \begin{equation}
                            \label{log-barrier-extension-problem}
                            \min_{\ttt} \quad {\cal E}(\ttt) + \sum_{i=1}^P\sum_{n=1}^{N} \tilde{\psi}_t \left ( f_i(s^n_{\ttt}) \right ) ,
                        \end{equation}
                        where $\tilde{\psi}_t$ is our {\em log-barrier extension}, which is convex, continuous and twice-differentiable:
                        \begin{equation}
                            \label{log-barrier extension}
                            \tilde{\psi}_{t}(z) =
                            \begin{cases}
                                    -\frac{1}{t} \log (-z) & \text{if } z \leq -\frac{1}{t^2}, \\
                                    tz - \frac{1}{t} \log (\frac{1}{t^2}) + \frac{1}{t} & \text{otherwise}.
                            \end{cases}
                        \end{equation}
                        \begin{figure}[t]
                                \centering
                                \begin{subfigure}[b]{0.48\columnwidth}
                                           \includegraphics[width=\textwidth]{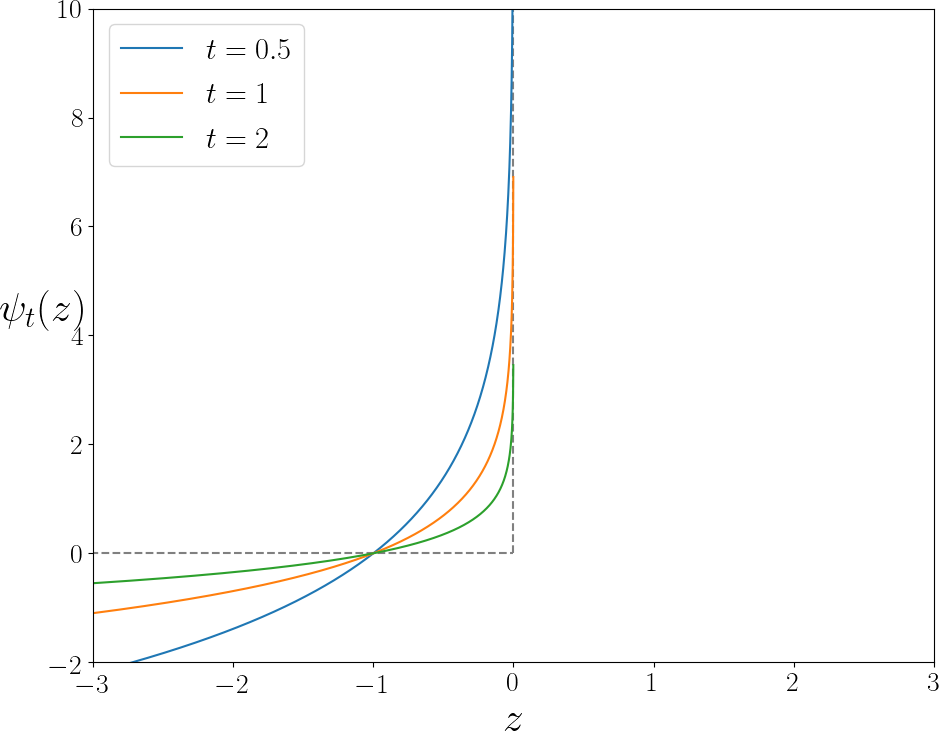}
                                           \caption{Standard log-barrier}
                                           \label{fig:logbarrier}
                                \end{subfigure}
                                \begin{subfigure}[b]{0.48\columnwidth}
                                           \includegraphics[width=\textwidth]{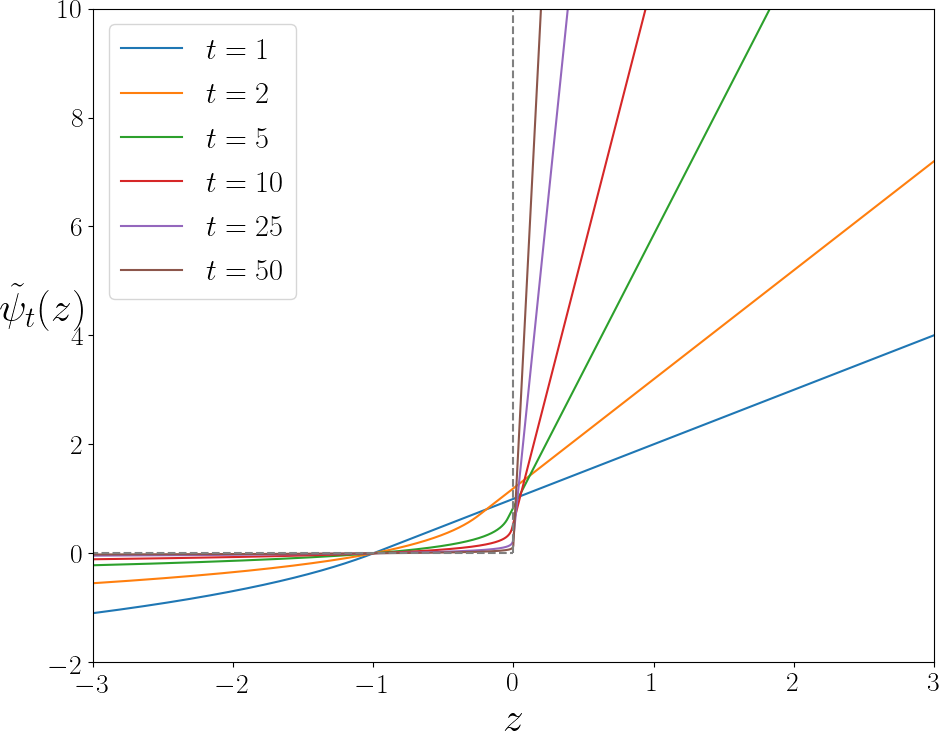}
                                           \caption{Log-barrier \emph{extension}}
                                           \label{fig:extlobarrier}
                                \end{subfigure}
                                \caption{Plots of the functions, for varying $t$.}
                        \end{figure}
                        The standard log-barrier $\psi_t$ and our proposed extension $\tilde \psi_t$ are depicted in sub-Figures \ref{fig:logbarrier} and \ref{fig:extlobarrier} respectively.
                        Similarly to the standard log-barrier, when $t \rightarrow + \infty$, our extension \eqref{log-barrier extension} can be viewed a smooth approximation of hard indicator function $H$. However, a very important difference is that the domain of our extension $\tilde{\psi}_{t}$ is not restricted to feasible points $\ttt$.
                        Therefore, our approximation \eqref{log-barrier-extension-problem} removes completely the requirement for
                        explicit Lagrangian-dual optimization for finding a feasible set of network parameters. In our case, the inequality constraints are fully handled within stochastic optimization, as in standard unconstrained losses, avoiding completely gradient ascent iterates and projections over {\em explicit} dual variables. 

                        In our approximation in Eq. \eqref{log-barrier-extension-problem}, the Lagrangian dual variables for the initial inequality-constrained problem of \eqref{Constrained-CNN}
                        are {\em implicit}. We show the following duality-gap bound, which yields sub-optimality certificates for feasible solutions of our approximation
                        in \eqref{log-barrier-extension-problem}. This result\footnote{The result applies to the general context of convex optimization. In deep CNNs, of course, a feasible solution of  our approximation may not be unique and is not guaranteed to be a global optimum as ${\cal E}$ and the constraints are not convex.} can be viewed as an extension of the standard result in \cite{Boyd2004}[page 566], which expresses the duality-gap as a function of $t$ for the log-barrier function.
                        \begin{Prop}
                            \label{prop:duality-gap-log-barrier-extension}
                            Let $\ttt^*$ be the solution of problem \eqref{log-barrier-extension-problem} and $\lbold^* \in \mathbb{R}^{P\times N}$ the corresponding vector of
                            implicit Lagrangian dual variables given by:
                            \begin{equation}
                                    \label{implicit-dual-barrier-extension-initial}
                                    \lambda^*_{i,n} = \begin{cases}
                                            -\frac{1}{t f_i(s^n_{\ttt^*})} & \text{if } f_i(s^n_{\ttt^*}) \leq -\frac{1}{t^2}, \\
                                            t & \text{otherwise.}
                                    \end{cases}
                            \end{equation}
                            Then, we have the following upper bound on the duality gap associated with primal $\ttt^*$ and implicit dual feasible $\lbold^*$ for the initial inequality-constrained problem \eqref{Constrained-CNN}:
                            \[ {\cal E}(\ttt^*) - g(\lbold^*) \leq PN/t. \]
                        \end{Prop}
                        \noindent{\em Proof:}

                        Let $\ttt^*$ be the solution of problem \eqref{log-barrier-extension-problem}
                        and $\lll^* \in \mathbb{R}^{P\times N}$ the corresponding vector of implicit dual variables given by \eqref{implicit-dual-barrier-extension-initial}.
                        We assume that $\ttt^*$ verifies approximately\footnote{When optimizing unconstrained loss via stochastic gradient descent (SGD), there is no guarantee that the obtained solution verifies exactly the optimality conditions.} the optimality condition for a minimum of \eqref{log-barrier-extension-problem}:
                        \begin{equation}
                                \label{optimality-condition-supp}
                                \nabla {\cal E}(\ttt^*) + \sum_{i=1}^P\sum_{n=1}^{N} \tilde{\psi'}_t \left ( f_i(s^n_{\ttt^*}) \right ) \nabla f_i(s^n_{\ttt^*}) \approx 0
                        \end{equation}
                        It is easy to verify that each dual variable $\lambda^*_{i,n}$ corresponds to the derivative of the log-barrier extension at $f_i(S_{\ttt^*})$:
                        \begin{equation*}
                                \label{eq:dummy}
                                \lambda^*_{i,n} = \tilde{\psi'}_t \left ( f_i(s^n_{\ttt^*}) \right )
                        \end{equation*}
                        Therefore, Eq. \eqref{optimality-condition-supp} means that
                        $\ttt^*$ verifies approximately the optimality condition for the Lagrangian corresponding to the original inequality-constrained problem in Eq. \eqref{Constrained-CNN} when $\lll = \lll^*$:
                        \begin{equation}
                                \label{eq:dummy2}
                                \nabla {\cal E}(\ttt^*) + \sum_{i=1}^P\sum_{n=1}^{N} \lambda^*_{i,n} \nabla f_i(s^n_{\ttt^*}) \approx 0
                        \end{equation}
                        It is also easy to check that the implicit dual variables defined in \eqref{implicit-dual-barrier-extension-initial} corresponds to a feasible dual, i.e., $\lll^*>0$
                        element-wise. Therefore, the dual function evaluated at $\lll^*>0$ is:
                        \[ g(\lll^*) = {\cal E}(\ttt^*) + \sum_{i=1}^P\sum_{n=1}^{N} \lambda_{i,n}^* f_i(s^n_{\ttt^*}), \]
                        which yields the duality gap associated with primal-dual pair $(\ttt^*, \lll^*)$:
                        \begin{equation}
                                \label{duality-gap-supp}
                                {\cal E}(\ttt^*) - g(\lll^*) = - \sum_{i=1}^P\sum_{n=1}^{N} \lambda_i^* f_i(s^n_{\ttt^*}).
                        \end{equation}
                        Now, to prove that this duality gap is upper-bounded by $PN/t$, we consider three cases for each term in the sum in \eqref{duality-gap-supp} and verify that, for all the cases, we have $\lambda_{i,n}^* f_i(s^n_{\ttt^*}) \geq -\frac{1}{t}$.
                        \begin{itemize}
                                \item $f_i(s^n_{\ttt^*}) \leq -\frac{1}{t^2}$: In this case, we can verify that $\lambda_{i,n}^* f_i(s^n_{\ttt^*}) = -\frac{1}{t}$ using the first line of \eqref{implicit-dual-barrier-extension-initial}.

                                \item $-\frac{1}{t^2} \leq f_i(s^n_{\ttt^*}) \leq 0$: In this case, we have $\lambda_{i,n}^* f_i(s^n_{\ttt^*}) = t f_i(s^n_{\ttt^*})$ from the second line of \eqref{implicit-dual-barrier-extension-initial}. As $t$ is strictly positive and $f_i(s^n_{\ttt^*}) \geq -\frac{1}{t^2}$, we have $t f_i(s^n_{\ttt^*}) \geq -\frac{1}{t}$, which means $\lambda_{i,n}^* f_i(s^n_{\ttt^*}) \geq -\frac{1}{t}$.

                                \item $ f_i(s^n_{\ttt^*}) \geq 0$: In this case, $\lambda_{i,n}^* f_i(s^n_{\ttt^*}) = t f_i(s^n_{\ttt^*}) \geq 0 >  -\frac{1}{t}$ because $t$ is strictly positive.
                        \end{itemize}

                        In all the three cases, we have $\lambda_{i,n}^* f_i(s^n_{\ttt^*}) \geq  -\frac{1}{t}$. Summing this inequality over $i$ gives:
                        \[ - \sum_{i=1}^P\sum_{n=1}^{N} \lambda_{i,n}^* f_i(s^n_{\ttt^*}) \leq  \frac{PN}{t} .\]
                        Using this inequality in \eqref{duality-gap-supp} yields the following upper bound on the duality gap associated with primal $\ttt^*$ and implicit dual feasible $\lll^*$ for the original inequality-constrained problem:
                        \[ {\cal E}(\ttt^*) - g(\lll^*) \leq PN/t. \]
                        \qed

                        This bound yields sub-optimality certificates for feasible solutions of our approximation in \eqref{log-barrier-extension-problem}. If the solution $\ttt^*$ that we obtain from our unconstrained problem \eqref{log-barrier-extension-problem} is feasible, i.e., it satisfies
                        constraints $f_i(s^n_{\ttt^*}) \leq 0$,
                        $\forall i, \forall n$,
                        then $\ttt^*$ is $PN/t$-suboptimal for the original inequality constrained problem: ${\cal E}(\ttt^*) - {\cal E}^* \leq PN/t$.
                        In deep CNNs, of course, a feasible solution for our approximation may not be unique and is not guaranteed to be a global optimum as ${\cal E}$ and the constraints are not convex.

                        From  Proposition \ref{prop:duality-gap-log-barrier-extension}, the following important fact follows immediately: If the solution $\ttt^*$ that we obtain from
                        unconstrained problem \eqref{log-barrier-extension-problem} is feasible and global, then it is $PN/t$-suboptimal for constrained problem \eqref{Constrained-CNN}:
                        ${\cal E}(\ttt^*) - {\cal E}^* \leq PN/t$.

                        Similarly to the standard log-barrier algorithm, we use a varying parameter $t$. At training time, we optimize a sequence of losses of the form \eqref{log-barrier-extension-problem} and increase gradually the value $t$ by a factor $\mu$. The network parameters obtained for the current $t$ and epoch are used as a starting point for the next $t$ and epoch. This effectively ``\emph{raises}'' the barrier over time.
                        We can summarize the fundamental differences between our log-barrier extension and a standard penalty function as follows:

                        A penalty does not act as a barrier near the boundary of the feasible set, i.e., a satisfied constraint yields null penalty and gradient. Therefore, at a given gradient update, there is nothing that prevents a satisfied constraint from being violated, causing oscillations between competing constraints and making the training unstable.
                        On the contrary, the strictly positive gradient of our log-barrier extension gets higher when a satisfied constraint approaches violation during optimization, pushing it back towards the feasible set.

                        Another fundamental difference is that the derivatives of our log-barrier extensions yield the implicit dual variables in Eq. \eqref{implicit-dual-barrier-extension-initial}, with sub-optimality and duality-gap guarantees, which is not the case for penalties. Therefore, our log-barrier extension mimics Lagrangian optimization, but with implicit rather than explicit
                        dual variables.

        \section{Experiments}
                Most of the existing methods---and the proposed log-barrier---are compatible with any differentiable function $f_i$, including non-linear and fractional terms, as in Eqs. \eqref{eq:soft_size} and \eqref{eq:soft_centroid} introduced further in the paper. However, we hypothesize that our log-barrier extension is better for handling the interplay between multiple competing constraints. To validate this hypothesis, we compare all strategies on the image segmentation tasks with constraints related to region size and location.
                As baselines we compare to a direct Lagrangian method, and a recent modification of the Lagrangian \cite{nandwani2019primal}.

                \paragraph{Region-size constraint}
                        We define the size (or volume) of a segmentation for class $k$ as the sum of its softmax predictions over the image domain:
                        \begin{equation}
                                \label{eq:soft_size}
                                \mathcal{V}_{k,\ttt}^n = \sum_{p \in \Omega} s^n_{k,p,\ttt} .
                        \end{equation}
                        We use the following inequality constraints on region size:
                        $0.9 \tau_{\mathcal{V}^n_k} \leq \mathcal{V}_{k,\ttt}^n \leq 1.1 \tau_{\mathcal{V}^n_k}$,
                        where, similarly to the experiments in \cite{kervadec2019constrained}, $\tau_{\mathcal{V}^n_k} = \sum_{p \in \Omega} y^n_{k,p}$ is determined from the ground truth $y^n$ of each image.

                \paragraph{Region-centroid constraints}
                        The centroid of the predicted region can be computed as a weighted average of the pixel coordinates:
                        \begin{equation}
                                \label{eq:soft_centroid}
                                \mathcal{C}_{k,\ttt}^n = \frac{\sum_{p\in\Omega} s_{k,p,\ttt}^n c_p}{\sum_{p \in \Omega} s_{k,p,\ttt}^n},
                        \end{equation}
                        where $c_p \in \mathbb{N}^2$ are the pixel coordinates on a 2D grid.
                        We constrain the position of the centroid in a box around the ground-truth centroid:
                        $ \tau_{\mathcal{C}^n_k} - 20 \leq \mathcal{C}_{k,\ttt}^n \leq \tau_{\mathcal{C}^n_k} + 20$,
                        with $\tau_{\mathcal{C}^n_k} = \frac{\sum_{p\in\Omega} y_{k,p}^n c_p}{\sum_{p \in \Omega} y_{k,p}^n}$ corresponding to the bound values associated
                        with each image.

                \paragraph{Bounding box tightness prior}
                        This prior \cite{hsu2019weakly,lempitsky2009image} assumes that any horizontal or vertical line inside the bounding box of an object of class $k$ will eventually cross the object. This can be generalized with segments of width $w$ inside the box, that will cross at least $w$ times the object. This prior can be easily reformulated as constraints. If $\mathcal{S}^n_L := \{s_l^n\}$ denotes the set of parallel segments to the sides of the bounding box for sample $n$, the following set of inequality constraints is trivial to define:
                        \begin{align}
                                \label{logbarrier:eq:tightness_constraints}
                                \sum_{p \in s^n_l} y^n_{k,p} &\geq w &\forall s^n_l \in \mathcal{S}^n_L, \forall n \in \mathcal{D} .
                        \end{align}
                        If we define the inside of the bounding box as $\Omega_F$, and the outside as $\Omega_B$ (such as $\Omega = \Omega_F \cup \Omega_B$ and $\Omega_F \cap \Omega_B = \{\emptyset\}$), we can define two other useful constraints for each image:
                        \begin{align}
                                \label{eq:emptyness_constraints} \sum_{p \in \Omega_B} s_{k,p,\ttt}^n &\leq 0 & \forall n \in \mathcal{D}, \\
                                \label{eq:global_size_reg} \sum_{p \in \Omega} s_{k,p,\ttt}^n &\leq |\Omega_F| & \forall n \in \mathcal{D}.
                        \end{align}
                        This setting is a good benchmark to evaluate the interplay of numerous, competing constraints simultaneously.

                \subsection{Datasets and evaluation metrics}
                        \label{ssec:dataset}
                        Experiments were performed on two different segmentation scenarios using synthetic and medical images.

                        \paragraph*{Synthetic images}
                        We randomly generated a synthetic dataset composed of 1100 images with two different circles of the same size but different intensity values, where the darker circle is the target region (Fig. \ref{fig:cherry_toy}, first column). Furthermore, different levels of Gaussian noise were added to the images. We employed 1000 images for training and 100 for validation.
                        We test the combinations of contraints \eqref{eq:soft_size} and \eqref{eq:soft_centroid}.

\begin{figure}
                                \centering
                                \includegraphics[width=.6\columnwidth]{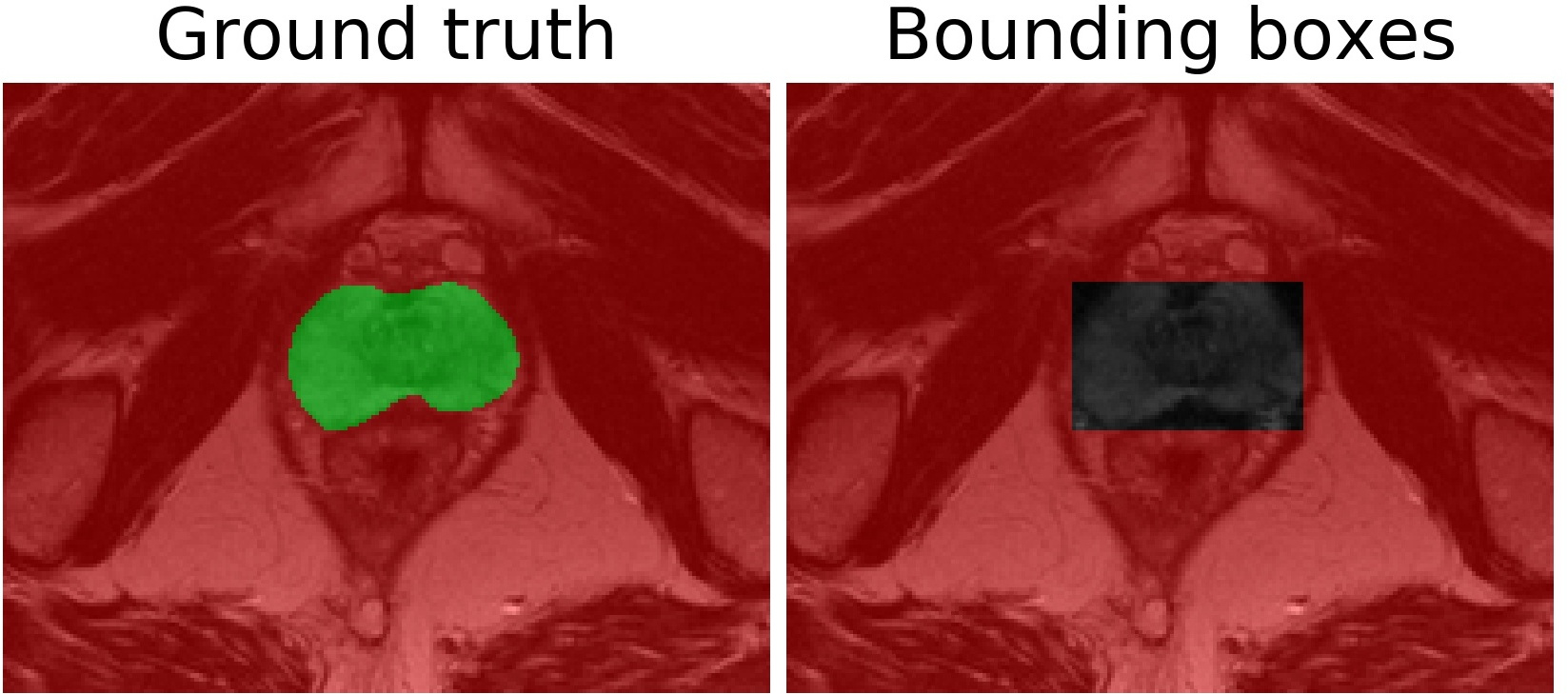}
                                \caption{Full mask of the prostate (\textit{left}) and the box annotations (\textit{right}). The background is in red and the foreground in green. No color means that no information is provided.}
                                \label{fig:prostate_labels}
                        \end{figure}

                        \paragraph*{Medical images} We use the dataset from the MICCAI 2012 prostate segmentation challenge \cite{litjens2014evaluation}, PROMISE12. This dataset contains Magnetic Resonance (MR) images from 50 patients, from which we employ 10 patients for validation and use the rest for training.
                        We test the combinations of constraints \eqref{logbarrier:eq:tightness_constraints}, \eqref{eq:emptyness_constraints} and \eqref{eq:global_size_reg}, with bounding boxes derived from the ground truth (illustrated in Figure \ref{fig:prostate_labels}).

                        \paragraph*{Evaluation} We resort to the common \text{Dice index} (DSC) = $\frac{2|S \bigcap Y|}{|S|+|Y|}$ to evaluate predicted segmentations. Furthermore, we evaluate the effectiveness and stability of the constrained optimization methods. 
                        To this end, we first compute at each epoch the percentage of constraints that are satisfied. Second, we measure the stability of the constraints, i.e., the percentage of constraints satisfied at epoch $t$ that are still satisfied at epoch $t+1$. And last, we measure the time needed to train a single epoch, including the dual update for the Standard Lagrangian and ReLU Lagrangian \cite{nandwani2019primal}.

                        The code is publicly available\footnote{\url{https://github.com/LIVIAETS/extended_logbarrier}}, and contains all relevant implementation details, hyper-parameters, and running scripts for easier reproducibility.

                \subsection{Results}
                        \label{sssec:results}

                        \paragraph{Quantitative results}
                                Results in terms of DSC are reported in Table \ref{tab:numbers}. The first thing we can observe on the synthetic dataset is that the standard Lagrangian, despite the introduction of a dedicated learning rate for its $\lbold$ update, is not able to learn when multiple constraints are in competition, i.e, DSC of 0.005 in the synthetic example. In addition, the ReLU Lagrangian approach proposed by \cite{nandwani2019primal} can better handle multiple constraints than a simple penalty \cite{He2017,kervadec2019constrained}.

                                With the high number of constraints and trivial solutions to balance, the proposed log-barrier extension learns successfully based on the information given by the constraints, compared to the other methods, achieving the best DSC across the two settings, and, more importantly, is the only method managing to predict non-empty segmentations on the medical task.

                                The very poor performance of penalty-based methods can be explained by the high-gradients generated when constraints are not satisfied, which leads to big and simplistic updates.

                                \begin{table}[t]
                                        \centering
                                        \caption{Mean DSC and standard deviation of the last 10 epochs on the validation on the toy example and PROMISE12 datasets.}
                                        \begin{tabular}{lccc}
                                                \toprule
                                                Method & Synthetic dataset & PROMISE12 \\
                                                \midrule
                                                Standard Lagrangian & 0.005 (0.014) & 0.000 (0.000) \\
                                                ReLU Lagrangian \cite{nandwani2019primal} & 0.798 (0.006) & 0.000 (0.000) \\
                                                \hline
                                                Penalty \cite{He2017,kervadec2019constrained} & 0.712 (0.022) & 0.000 (0.000) \\
                                                Log-barrier extensions (ours) & \textbf{0.945} (0.001) & \textbf{0.813} (0.024) &  \\
                                                \hline
                                                Full supervision & 0.998 (0.000) & 0.880 (0.001) \\
                                                \bottomrule
                                        \end{tabular}
                                        \label{tab:numbers}
                                \end{table}

                        \begin{figure}[ht]
                                \centering
                                \includegraphics[width=\columnwidth]{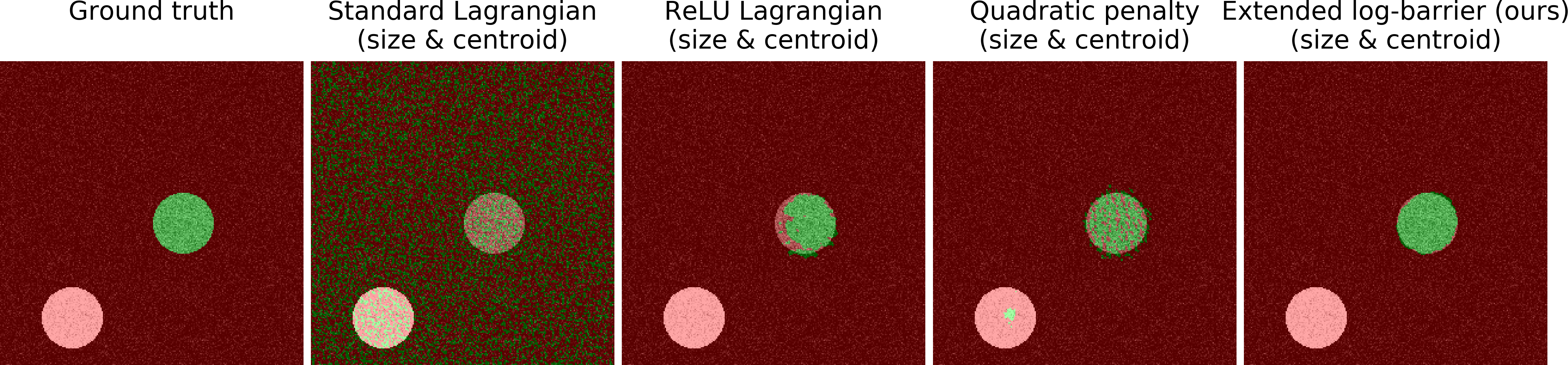}
                                \caption{Results on the synthetic dataset (background in red and foreground in green).}
                                \label{fig:cherry_toy}
                        \end{figure}

                        \paragraph{Qualitative results}
                                A visual comparison on the synthetic dataset is depicted in Figure \ref{fig:cherry_toy}. In this figure we can first observe that standard Lagrangian generates noisy segmentations, which is in line with the quantitative results reported in Table \ref{tab:numbers}. Both ReLU Lagrangian \cite{nandwani2019primal} and penalty-based methods obtain better target segmentations. Nevertheless, they cannot handle efficiently the interplay between multiple constraints.
                                Meanwhile, the proposed extended log-barrier demonstrates a strong ability to handle several constraints simultaneously, which is reflected in the circle segmentation close to the ground truth.

\begin{figure}[ht]
                                        \centering
                                        \begin{subfigure}[b]{0.31\columnwidth}
                                                   \includegraphics[width=\textwidth]{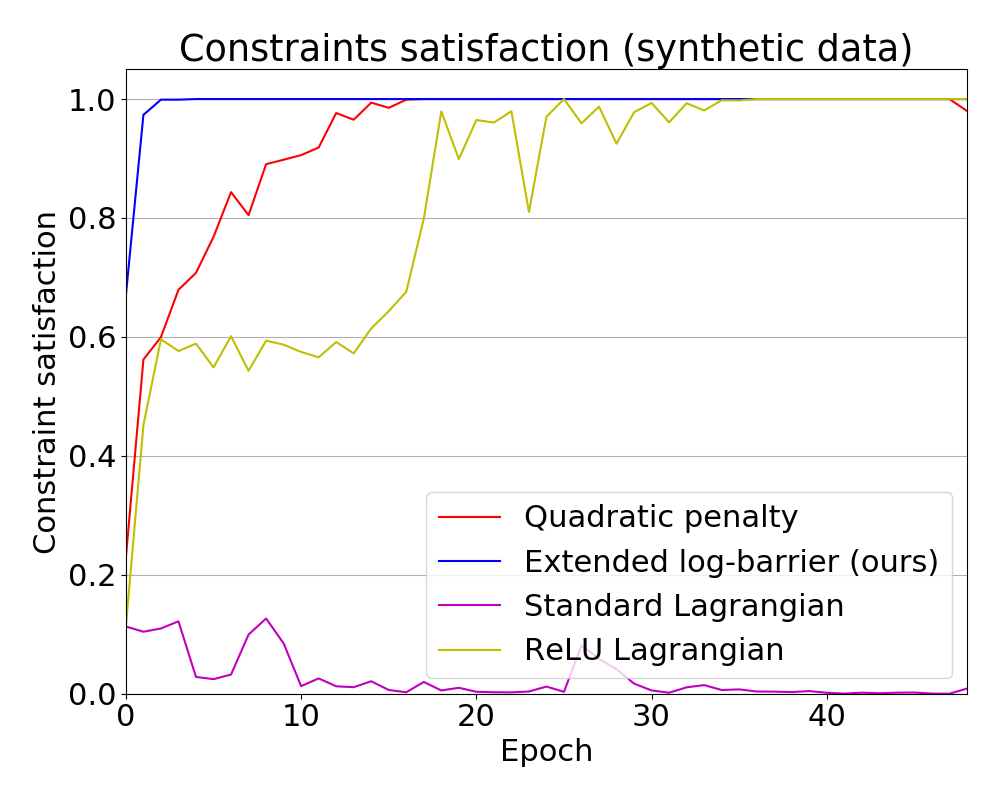}
                                        \end{subfigure}
                                        \begin{subfigure}[b]{0.31\columnwidth}
                                           \includegraphics[width=\textwidth]{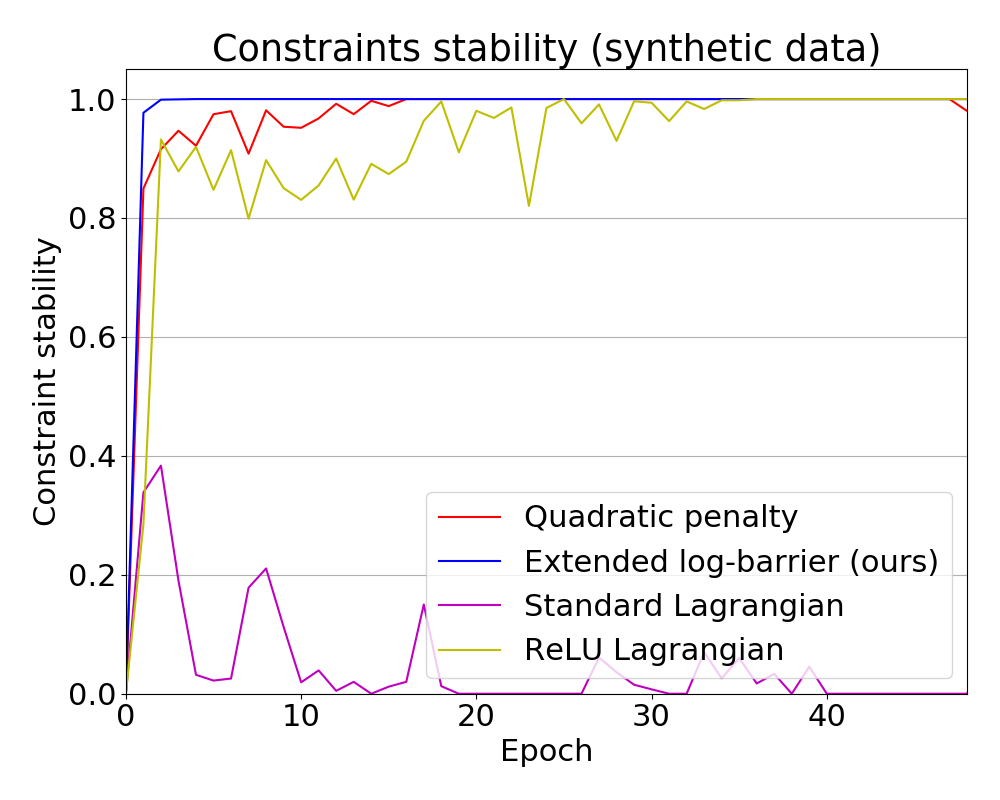}
                                        \end{subfigure}
                                        \begin{subfigure}[b]{0.31\columnwidth}
                                           \includegraphics[width=\textwidth]{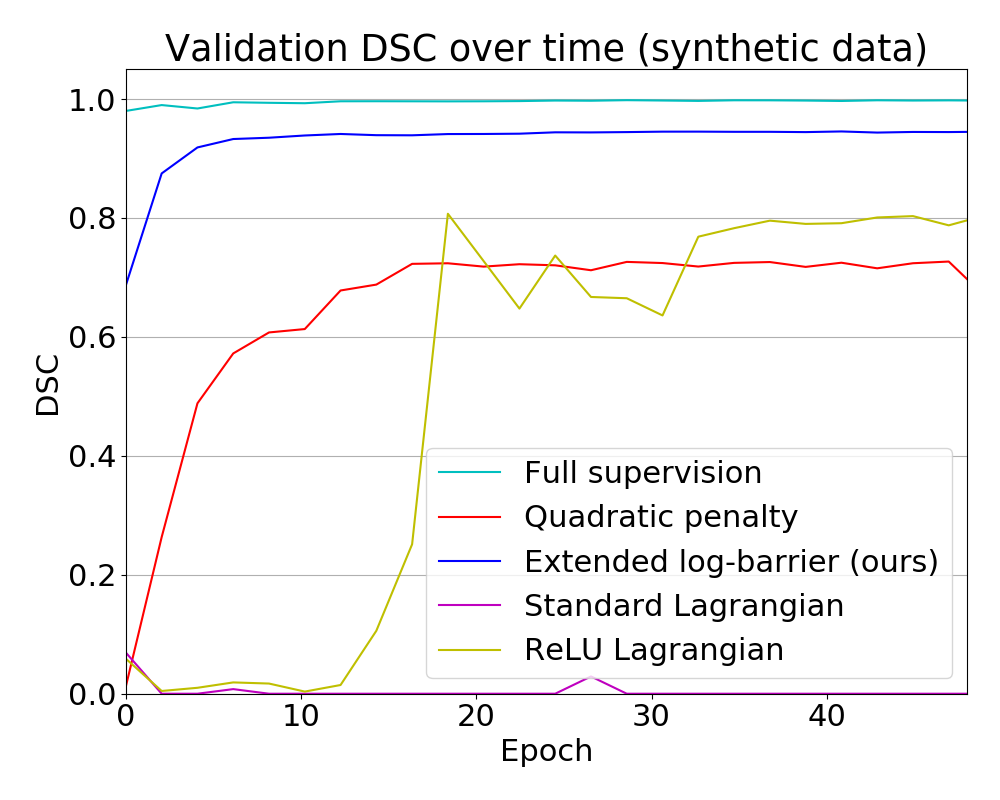}
                                        \end{subfigure}
                                        \\
                                        \begin{subfigure}[b]{0.31\columnwidth}
                                           \includegraphics[width=\textwidth]{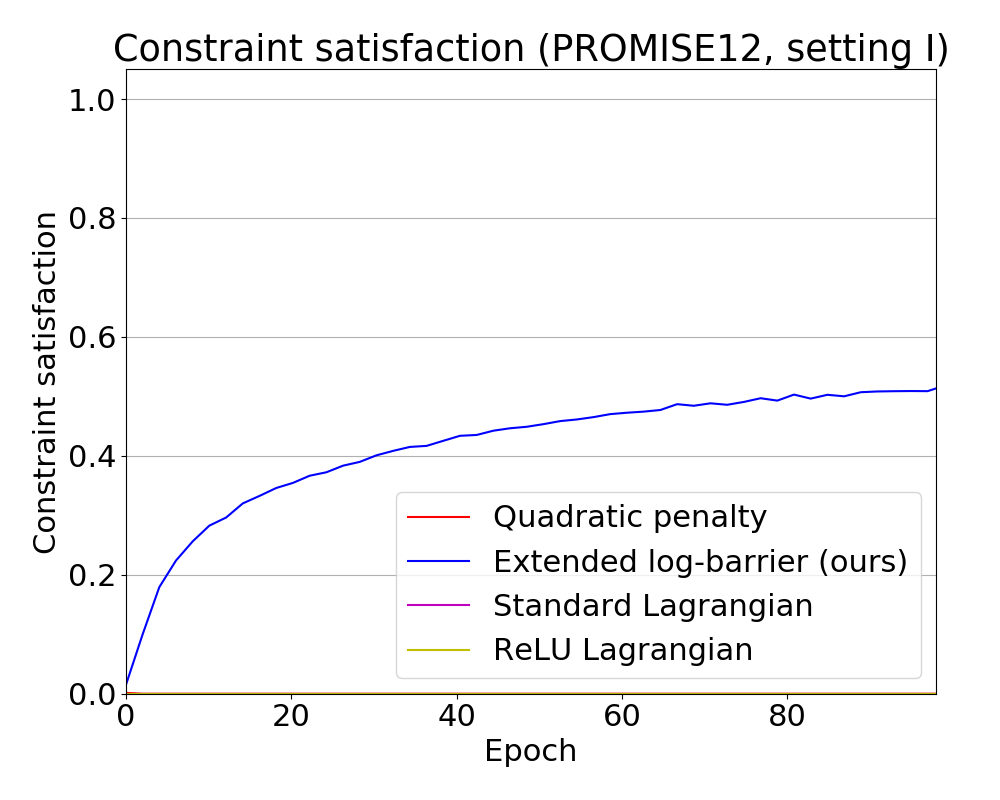}
                                        \end{subfigure}
                                        \begin{subfigure}[b]{0.31\columnwidth}
                                          \includegraphics[width=\textwidth]{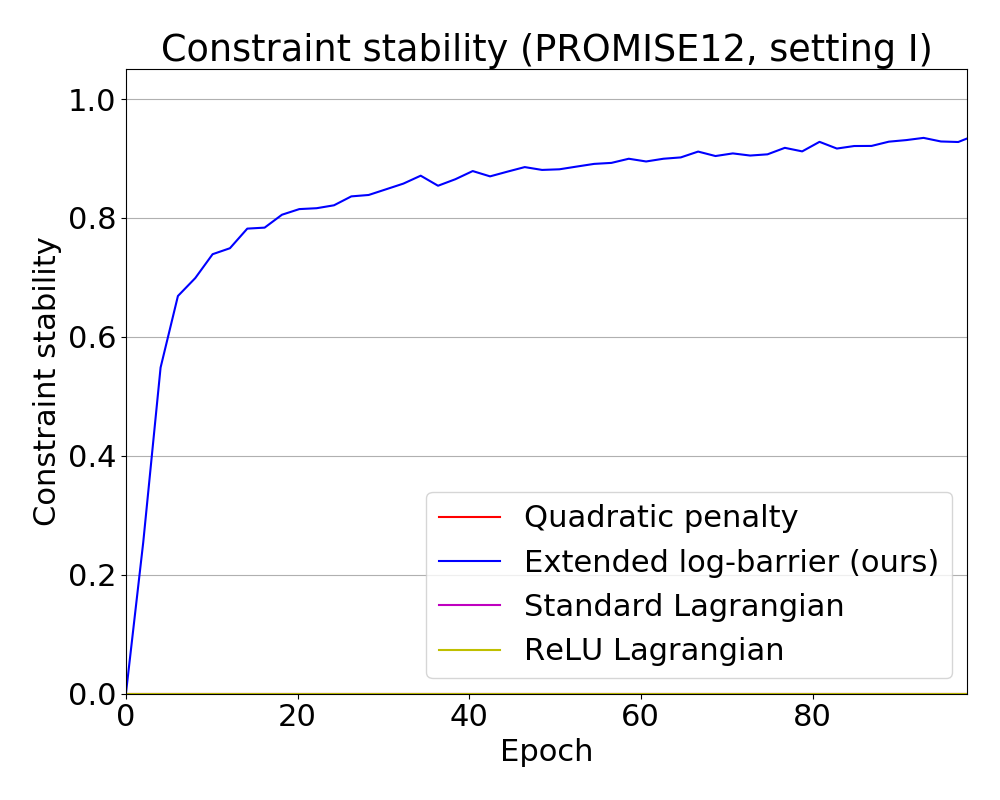}
                                        \end{subfigure}
                                        \begin{subfigure}[b]{0.31\columnwidth}
                                            \includegraphics[width=\textwidth]{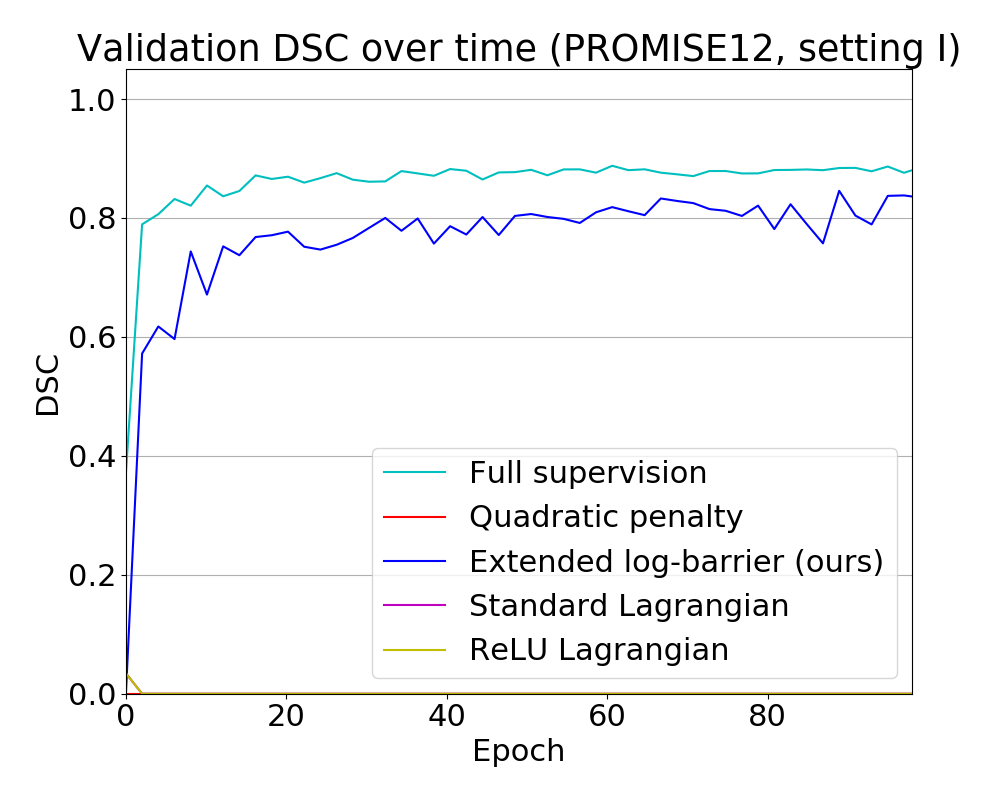}
                                        \end{subfigure}
                                        \caption{Constraints satisfaction, stability and DSC evolution on different settings. Best viewer in colors.}
                                        \label{logbarrier:fig:learning_curves}
                                \end{figure}

                        \paragraph{Constraints satisfaction and stability}
                                The metrics over training epochs are shown in Fig. \ref{logbarrier:fig:learning_curves}.
                                We can notice that on top of the better absolute performances, the proposed log-barrier extension is also more stable during training, both in performance and constraints satisfaction.

                        \paragraph{Computational cost and efficiency}
                                Penalties and the proposed log-barrier extension have negligible cost compared to optimizing the base-loss $\mathcal{E}(\ttt)$ alone (up to 5\% slowdown when the number of constraints becomes very high). In contrast, Lagrangian methods incur in higher computational cost. For example, in the standard and ReLU Lagrangian, it amounts to nearly a 25\% slowdown (due to the extra loop over the training set to perform the $\lbold$ update).

        \section{Conclusion}
                We proposed \emph{log-barrier extensions}, which approximate Lagrangian optimization of constrained-CNN problems with a sequence of unconstrained losses. Our formulation relaxes the need for an initial feasible solution, unlike standard interior-point and log-barrier methods. This makes it convenient for deep networks. We also provided an upper bound on the duality gap for our proposed extensions, thereby generalizing the duality-gap result of standard log-barriers and showing that our formulation has dual variables that mimic implicitly (without dual projections/steps) Lagrangian optimization.
                Therefore, our implicit Lagrangian formulation can be fully handled with SGD, the workhorse of deep networks.
                We reported constrained-CNN experiments, showing that log-barrier extensions outperform several other types of Lagrangian methods and penalties, in terms of accuracy and training stability.
                Log-barrier extensions can be useful in breadth of problems in vision and learning, where constraints occur naturally. This include, for instance, adversarial robustness \cite{Rony2019CVPR}, stabilizing the training of GANs \cite{GulrajaniNIPS17}, domain adaptation for segmentation \cite{curriculumDA2019}, pose-constrained image generation \cite{poseconstraintsneurips2018},
                3D human pose estimation \cite{Marquez-Neila2017}, deep reinforcement learning \cite{He2017} and natural language processing \cite{nandwani2019primal}. To our knowledge, those constraints are typically handled with basic penalties; it will therefore be interesting to investigate log-barrier extensions in these diverse contexts.

        \bibliographystyle{plain}
        \bibliography{biblio}
\end{document}